\documentclass{article}
\usepackage{spconf,amsmath,graphicx}
\usepackage{mathtools}
\usepackage{multirow}
\usepackage{hhline}
\usepackage{arydshln}

\DeclarePairedDelimiter\abs{\lvert}{\rvert}%
\DeclarePairedDelimiter\norm{\lVert}{\rVert}%

\title{MULTI-SCENARIO DEEP LEARNING FOR MULTI-SPEAKER SOURCE SEPARATION}
\name{Jeroen Zegers, Hugo Van hamme}
\address{KU Leuven, Dept. ESAT, Belgium}
\begin{document}
%
\maketitle
\begin{abstract}
Research in deep learning for multi-speaker source separation has received a boost in the last years. However, most studies are restricted to mixtures of a specific number of speakers, called a specific scenario. While some works included experiments for different scenarios, research towards combining data of different scenarios or creating a single model for multiple scenarios have been very rare. In this work it is shown that data of a specific scenario is relevant for solving another scenario. Furthermore, it is concluded that a single model, trained on different scenarios is capable of matching performance of scenario specific models.
\end{abstract}
\section{Introduction}
\label{sec:intro}
Source separation (SS) in audio refers to the task of retrieving the original sound signal of multiple sound source objects that have been recorded in a mixture. More specifically, this task focuses on the case when the sound sources are (partially) active at the same time. Source separation can be split in two domains: speech vs noise separation, also called speech enhancement (SE), and speech vs speech separation, also called multi-speaker source separation (MSSS). In recent years there have been many studies in deep learning approaches for SS. Examples in SE are \cite{xu2014experimental,weninger2014single,wang2013towards,Zhang:2016:DEL:2992480.2992491,adilouglu2016variational} and for MSSS there have been studies for male - female speech separation \cite{huang2014deep} and speaker dependent source separation \cite{huang2015joint}. All these works handle an interclass separation problem since distinctive classes can be defined: speech and noise; male and female; Alex and Bob. In unsupervised (gender independent) MSSS, no assumptions on the sources can been made and only a single class can be defined, namely a speaker class. 
This makes unsupervised MSSS an intraclass separation problem, which is intrinsically harder than the other discussed problems, as will be explained in section \ref{sec:task}. This paper handles single microphone problems to be as general as possible.

Different approaches to solve this unsupervised MSSS problem have been proposed, all of which try to cope with the permutation problem discussed in section \ref{sec:task}. In Deep Clustering (DC), every time-frequency bin of a mixture is mapped to an embedding vector and these are then clustered per speaker using K-means (see section \ref{sec:DC}) \cite{hershey2016deep,isik2016single}. Another approach is to directly estimate each source signal and use utterance-level Permutation Invariant Training (uPIT) to cope with the permutation problem (see section \ref{sec:uPIT}) \cite{Yu2016permutation}. Deep Attractor Nets (DANet) are a combination of DC and uPIT as they also estimate embeddings, but the clustering is done in the netwerk itself via attractors, so that source signals can be estimated directly \cite{chen2017deep}. In \cite{DBLP:journals/corr/StephensonCGN17} embeddings are being pulled to a corresponding speaker vector during training and both are jointly optimized. Finally, in \cite{zegers2017improving,zmolikova2017speaker} approaches have been presented to include speaker information in an (un)supervised manner.

In section \ref{sec:MSSS} two methods will be explained on how to train and test a model on a specific scenario for a mixture of $S$ speakers. If a solution is requested for multiple scenarios (e.g. both mixtures of $S_1$ and $S_2$ speakers), a simple solution would be to train separate models for each scenario. However, this has the disadvantage that a model can only be trained on data for a specific scenario, while data on other scenarios could still be relevant. Furthermore, the solution would contain as many models as there are scenarios and for each presented mixture, the requested model would have to be selected. 

Therefore a solution for multi-scenario learning is proposed for both methods. This allows the model to learn from more data and only a single model has to be retained for all possible scenarios. In \cite{isik2016single} a small initial experiment in this context has already been done.

The remainder of this paper is organized as follows. In section \ref{sec:MSSS} a brief overview of MSSS and the permutation problem is given, as well as an explanation of two methods that have previously been used for the task. In section \ref{sec:MSL} a solution for multi-scenario learning is given. Experiments are presented in section \ref{sec:exp} and a conclusion is given in section \ref{sec:conc}.

\section{Multi-speaker source separation}
\label{sec:MSSS}

\subsection{Task and permutation problem}
\label{sec:task}
In the tasks of MSSS, one wants to estimate a signal $\hat{\mathbf{x}}_s[n]$ for the $s^{th}$ speaker that is as close as possible to the source signal $\mathbf{x}_s[n]$, given a mixture signal $\mathbf{y}[n]=\sum_{s=1}^{S}\mathbf{x}_s[n]$ of $S$ speakers. In the time-frequency domain, the same task can be expressed using the Short Time Fourier Transform (STFT) of the signals. The task is then to estimate $\hat{\mathbf{X}}_s(t,f)$ from $\mathbf{Y}(t,f)=\sum_{s=1}^{S}\mathbf{X}_s(t,f)$. Usually, a mask $\hat{\mathbf{M}}_s(t,f)$ is estimated for the $s^{th}$ speaker such that 
\begin{equation}
\label{eq:spec_est}
\hat{\mathbf{X}}_s(t,f)=\hat{\mathbf{M}}_s(t,f)\circ\mathbf{Y}(t,f)
\end{equation}
for every time frame $t=0, \ldots, T$ and every frequency $f=0, \ldots, F$ and with $\circ$ the Hadamard product \cite{Yu2016permutation}. The task to solve the MSSS problem is thus to find a mapping $f_{\theta,s}$ from the input mixture to the mask estimates.
\begin{equation}
\label{eq:mask_est}
\hat{\mathbf{M}}_s(t,f)=f_{\theta,s}(\mathbf{Y})
\end{equation}
with the constraint that $\hat{\mathbf{M}}_s(t,f)\ge 0$ and $\sum_{s=1}^{S}\hat{\mathbf{M}}_s(t,f)=1$ for every time-frequency bin (t,f).

A straightforward approach is to use a deep learning model for the mapping and train the parameters $\theta$ to minimize a loss function 
\begin{equation}
\label{eq:general_loss}
\mathcal{L}_{\theta} = \sum_{s=1}^{S}\sum_{t,f} D(\abs{\hat{\mathbf{X}}_{\theta,s}(t,f)},\abs{\mathbf{X}_s(t,f)})
\end{equation}
with $D$ some distance measure. However, since an intraclass separation task is executed and no prior information on the speakers is assumed to be known, there is no guarantee that the network is consistent in its assignment of speakers. This is the label ambiguity or permutation problem. To cope with this ambiguity, a loss function has to be defined that is independent of the order of the speaker targets. Deep Clustering (DC) and utterance-level Permutation Invariant Training (uPIT), both use such a permutation invariant loss function.

\subsection{Deep Clustering}
\label{sec:DC}
In DC, the masks are not estimated directly from the input mixture, as was done in equation \ref{eq:mask_est}. Instead a $D$-dimensional embedding vector $\mathbf{v}_{tf}$ is found for every time-frequency bin via a mapping $\mathbf{v}_{tf}=f_{\theta}(\mathbf{Y})$. $f_{\theta}$ is chosen such that $\mathbf{v}_{tf}$ is normalized to unit length. The embedding vectors for every time-frequency bin are stored as rows in a ($TF\times D$)-dimensional matrix $\mathbf{V}$. Define a ($TF\times S$)-dimensional target matrix $\mathbf{W}$, so that $w_{tf,s}=1$ if target speaker $s$ has the most energy in the mixture for bin $(t,f)$ and $w_{tf,s}=0$ otherwise. A permutation independent loss function (the columns in $\mathbf{W}$ can be interchanged without changing the loss function) is then stated as
\begin{equation}
\label{eq:dc_loss}
\begin{split}
\mathcal{L}_{\theta} &= \norm{\mathbf{V}\mathbf{V}^T-\mathbf{W}\mathbf{W}^T}_F^2 \\
&= \sum_{t_1,f_1,t_2,f_2}(\langle \mathbf{v}_{t_1f_1},\mathbf{v}_{t_2f_2}\rangle-\langle \mathbf{w}_{t_1f_1},\mathbf{w}_{t_2f_2}\rangle)^2
\end{split}
\end{equation}
where $\norm{.}_F^2$ is the squared Frobenius norm. Since $\mathbf{w}_{tf}$ is a one-hot vector,
\begin{equation}
\langle \mathbf{w}_{t_1f_1},\mathbf{w}_{t_2f_2}\rangle=
\begin{cases}
      1, & \text{if}\ \mathbf{w}_{t_1f_1}=\mathbf{w}_{t_2f_2} \\
      0, & \text{otherwise}
    \end{cases}.
\end{equation}
The angle $\phi_{t_1f_1,t_2f_2}$ between the normalized vectors $\mathbf{v}_{t_1f_1}$ and $\mathbf{v}_{t_2f_2}$ is thus ideally 
\begin{equation}
\phi_{t_1f_1,t_2f_2}=
\begin{cases}
      0, & \text{if}\ \mathbf{w}_{t_1f_1}=\mathbf{w}_{t_2f_2} \\
      \pi/2, & \text{otherwise}
    \end{cases}.
\end{equation}
Afterwards, all embedding vectors are clustered into $S$ clusters $c$ using K-means. The masks are then constructed as follows
\begin{equation}
\hat{\mathbf{M}}_{s,tf}=
\begin{cases}
      1, & \text{if}\ \mathbf{v}_{tf} \in c_s \\
      0, & \text{otherwise}
    \end{cases}.
\end{equation}
Equation \ref{eq:spec_est} can then be used to estimate the original source signals.

The network architecture is independent of the number of speakers present in the mixture (the number of output nodes is dependent on the embedding dimension $D$, which is independent of $S$) and can thus be used for a mixture of any number of speakers. However, for the K-means clustering, the number of clusters has to be known. Initial experiments using X-means with the Bayesian Information Criterion, showed that automatically finding the number of clusters (or speakers) in the embedding space is far from evident.

\subsection{Utterance-level Permutation Invariant Training}
\label{sec:uPIT}
uPIT has a loss function based on equation \ref{eq:general_loss}, but it has been adjusted to cope with the label ambiguity. The loss is defined as
\begin{equation}
\label{eq:pit_loss}
\mathcal{L}_{\theta} = \min_{p\,\in\,\mathcal P_S} \sum_{s=1}^{S}\sum_{t,f} \norm{\abs{\hat{\mathbf{X}}_{\theta,s}(t,f)}-\abs{\mathbf{X}_{p_s}(t,f)}}_F^2 
\end{equation}
with $\mathcal P_S$ the set of all possible permutations for $S$ members and $\hat{\mathbf{X}}_{\theta,s}(t,f)$ is as defined in equation \ref{eq:spec_est} and \ref{eq:mask_est}.

Since every mask requires its own output nodes, the network is dependent on the number of speakers. The number of permutations to check in equation \ref{eq:pit_loss} is $S!$, but can be implemented to have a computational complexity of $S^2$. 
This could still give computational problems for high $S$, but that seems unrealistic for the task of MSSS.

\section{Multi-scenario learning}
\label{sec:MSL}
In the introduction some benefits of multi-scenario learning were discussed. The DC algorithm is suited the most for multi-scenario learning, since the architecture of the network can be chosen identical for any scenario. For uPIT, the network is shared, except for the output layer, which is specific for each scenario. 


Generally, in single scenario learning a loss $\mathcal{L}(\theta,\mathbf{I},\mathbf{T})$ is defined, based on the model inputs $\mathbf{I}$, the model parameters $\theta$ and the model targets $\mathbf{T}$. After each training batch each parameter is updated as follows
\begin{equation}
\label{delta_theta}
\Delta\theta_i=g\left(\frac{\partial{\mathcal{L}(\theta,\mathbf{I},\mathbf{T})}}{\partial{\theta_i}}\right)
\end{equation}
where $g()$ is specific to the learning algorithm, e.g. stochastic gradient descent or Adam, and $\mathcal{L}$ can refer to equation \ref{eq:dc_loss} or \ref{eq:pit_loss}.

For multi-scenario learning, consisting of $J$ scenarios, a solution would be to define a joint loss as
\begin{equation}
\label{eq:joint_loss}
\mathcal{L}_{m}(\theta,\mathbf{I}_{m},\mathbf{T}_{m}) = \sum_{j=1}^{J} \alpha_{j} \mathcal{L}_{j}(\theta,\mathbf{I}_{j},\mathbf{T}_{j})
\end{equation}
and take $\alpha_1=1$, with $\mathcal{L}_{j}(\theta,\mathbf{I}_{j},\mathbf{T}_{j})$ a single scenario loss. 
In the remainder of this paper $\mathcal{L}_{j}(\theta,\mathbf{I}_{j},\mathbf{T}_{j})$ is referred to as $\mathcal{L}_{j}$, to simplify notations. Equation  \ref{delta_theta} can then be applied as usual. However, $J-1$ meta parameters $\alpha$ have to be chosen, e.g. by tuning them using a validation set. To circumvent this we define a parameter update for each scenario and sum over these updates as follows
\begin{equation}
\label{eq:acc_delta}
\begin{split}
\Delta_{m}\theta_i = \sum_{j}^{J} \Delta_{j}\theta_i &= \sum_{j}^{J} g\left(\frac{\partial{\alpha_{j} \mathcal{L}_{j}}}{\partial{\theta_i}}\right) \\
&= \sum_{j}^{J} g\left(\frac{\alpha_{j}\partial{ \mathcal{L}_{j}}}{\partial{\theta_i}}\right)
\end{split}
\end{equation}
where equation \ref{delta_theta} and \ref{eq:joint_loss} were used in the second step. If a learning algorithm is chosen that is independent of the scale of the gradients, or in other words $g(\alpha\partial{\mathcal{L}}/\partial{\theta_i})=g(\partial{\mathcal{L}}/\partial{\theta_i)}$, like the Adam algorithm \cite{kingma2014adam}, equation \ref{eq:acc_delta} simplifies to 
\begin{equation}
\label{eq:acc_delta_simple}
\Delta_{m}\theta_i = \sum_{j}^{J} g\left(\frac{\partial{ \mathcal{L}_{j}}}{\partial{\theta_i}}\right)
\end{equation}
which is independent of the loss scale factors $\alpha$ so that they no longer have to be tuned. To clarify, if in equation \ref{eq:joint_loss} the loss of a specific scenario would be very dominant over the others, the other scenarios would almost be ignored for the parameter updates (the $\alpha$ parameters are introduced to counter this). However, if instead, each scenario calculates a parameter update using the Adam algorithm, the relative magnitudes of the losses between scenarios no longer play a role due to the invariance to the magnitude of the loss for Adam. Notice, if a learning algorithm like stochastic gradient descent would be chosen, the parameter updates would scale to the magnitude of the losses and the global update would be the same as using equation \ref{eq:joint_loss}.  A possible downside of the chosen strategy is that relative importance of the scenarios can no longer be tuned.
For DC, the scenario specific gradients $\partial{\mathcal{L}_{j}}/{\partial{\theta_i}}$ are defined for every scenario for every parameter $\theta_i$. In uPIT, the gradient for a scenario is set to 0 for parameters of the output layer that correspond to a different scenario.

\section{Experiments}
\label{sec:exp}

\subsection{Experimental setup}
All experiments were done using the corpus introduced in \cite{hershey2016deep}, which contains artificial mixtures created by mixing together single speaker utterances from the Wall Street Journal 0 (WSJ0) corpus. For every utterance a gain factor was randomly chosen between 0 and 5 dB and utterances were sampled at 8kHz. The length of the mixture was chosen equal to the shortest utterance in the mixture as to maximize the overlap.
The training and validation sets contained 20,000 and 5,000 mixtures, respectively and where taken from the \texttt{si\_tr\_s} set of WSJ0. The test set contained 3,000 mixtures using 16 held-out speakers of the \texttt{si\_dt\_05} and \texttt{si\_et\_05} set. The log-magnitude (floored at -300) of the STFT with a 32ms window length and 8ms hop size were used as features and where normalized with mean and variance, calculated over the whole training set.

For both methods the deep learning network has 2 fully connected bidirectional long short-term memory (BLSTM) layers with 600 hidden units each, using a tanh activation \cite{hochreiter1997long}. The parameters of these models where updated using the Adam learning algorithm with initial learning rate $10^{-3}$, $\beta_1=0.9$, $\beta_2=0.999$ and $\epsilon=10^{-8}$. For DC the embedding dimension was chosen at $D=20$ and since the frequency dimension was $F=129$, the total number of output nodes was $DF=20*129=2580$. For uPIT there are $SF=S*129$ output nodes. Zero mean Gaussian noise with standard deviation 0.2 was applied to the training features. Dropout was not used since it did not improve the results in the experiments. Early stopping was applied when the validation loss increased for 4 consecutive times. The networks were trained using curriculum learning \cite{bengio2009curriculum}, i.e. the networks were presented an easier task before tackling the main task. Here, the network was first trained on 100-frame non-overlapping segments of the mixtures. This network was then used as initialization when training over the full mixture. Performance for MSSS was measured in signal-to-distortion ratio (SDR) improvements on the test set, using the \texttt{bss\_eval} toolbox \cite{vincent2006performance}. All networks were trained using TensorFlow \cite{abadi2016tensorflow} and the code for all the experiments can be found here: \\\texttt{https://github.com/JeroenZegers/Nabu-MSSS}. Results for all experiments are summarized in table \ref{tab:res} for same gender mixtures (SG), mixtures with both genders (BG) and all mixtures (av). Only results on all mixtures (av) will be discussed.


\begin{table*}[t]
  \centering
  \begin{tabular}{c c c | c c c : c c c : c c c }
    algo- & train & valid. & \multicolumn{9}{c}{test set} \\
    rithm & set & set & \multicolumn{3}{c}{2spk} & \multicolumn{3}{:c}{3spk} & \multicolumn{3}{:c}{2+3spk}  \\
     &  &  & SG & BG & av & SG & BG & av  & SG & BG & av  \\
    \hline
    \multirow{6}{*}{DC} & \multirow{2}{*}{2spk} & 2spk & 6.69 & 10.71 & \textbf{8.73} & 1.16 & 2.21 & 1.96  & 4.85 & 5.61 & 5.35  \\
     &  & 3spk & - & - & - & 1.23 & 2.18 & 1.96  &- & - &  -  \\
     & \multirow{2}{*}{3spk} & 2spk & 5.67 & 10.06 & 7.89 & - & - & - & - &  -& -   \\
     &  & 3spk & 6.05 & 10.26 & 8.18 & 3.39 & 6.69 & \textbf{5.91}  & 5.16 & 8.12 & 6.67  \\
     & 2+3spk & 2+3spk & 6.54 & 10.22 & 8.41 & 3.40 & 6.13 & 5.48  & 5.49 & 7.81 & \textbf{6.95}  \\
     & 2+3spk half & 2+3spk & 5.93 & 10.07 & 8.03 & 2.81 & 5.83 & 5.11 & 4.89 & 7.53 & 6.57   \\
     \hline   
    \multirow{4}{*}{uPIT} & 2spk & 2spk & 5.85 & 10.20 & 8.05 & - & - & -  & - & - & -   \\
     & 3spk & 3spk & -  & - & - & 3.94 & 7.09 & 6.34 & - & - &  -  \\
     & 2+3spk & 2+3spk & 6.43 & 10.58 & \textbf{8.53} & 4.18 & 7.23 & \textbf{6.50}  & 5.68 & 8.57 & \textbf{7.52}   \\
     & 2+3spk half & 2+3spk & 5.90 & 10.27 & 8.11 & 3.91 & 6.86 & 6.16  & 5.24 & 8.22 & 7.14  

  \end{tabular}
  \caption{SDR improvement results for DC and uPIT for different train/validation/test sets. Results are shown for same gender mixtures (SG), mixtures with both genders (BG) and all mixtures (av). When mentioned, only half of the training set is used. Entries with '-' refer to a result that was not plausible or was deemed not relevant.}
  \label{tab:res}
\end{table*}

\subsection{Single scenario learning}
First, some single scenario learning experiments are performed using DC. Models are trained on 2 or 3 speakers and testing is done on 2 and 3 speakers. It is noticed that testing on 3 speakers when only seen mixtures of 2 speakers during training, drastically degrades performance compared to training on 3 mixtures (-3.95 dB). However, for the reversed scenario, the drop in performance is much lower (-0.55 dB). Being able to separate 3 speakers seems to partly rely on the subtask of separating 2 speakers.

In the next experiment, the validation and testing scenario differed from the training scenario. This is compared with the experiment where the testing scenario differed from the training and validation scenario. This was done to see whether a network trained on a specific single scenario would be able to generalize better to a different scenario, should it not be over-trained to the specific scenario. No significant difference was found however, so the conclusion from the previous paragraph is retained.

Finally, the performance of uPIT and DC are compared (only for within scenario, since uPIT cannot directly be used out of scenario) and it is concluded that DC outperforms uPIT for 2 speaker mixtures and uPIT is better for 3 speaker mixtures. Differences are rather small and possibly dependent on the chosen network architecture.

\subsection{Multi-scenario learning}


For the multi-scenario learning experiments (train 2+3spk), DC falls just short in comparison with the best solution for 2 speaker mixtures (-0.32 dB) and 3 speaker mixtures (-0.43 dB). However, for uPIT the multi-scenario learning experiment slightly outperforms the best solution for both 2 speakers (+0.48 dB) and 3 speakers (+0.16 dB). Not only has the model succeeded to use training data from a scenario different from test scenario, it has also managed to create a single model for both test scenarios. The latter is emphasized to make a distinction with the experiment where the BLSTM layers would be fine tuned to the specific test scenario so possibly better performance could be achieved, but this would lead again to the need of a model per test scenario (even though data from different scenarios has been used). A reason that uPIT is better suited for multi-scenario learning than DC could be that while most parameters are shared over the scenarios, the output layer is still allowed to be scenario specific. In following work the performance of DC for multi-scenario learning with a scenario specific output layer will be evaluated.

If only half of the training sets of both scenarios are used, the total number of data samples for multi-scenario learning remains the same compared to the single scenario cases. 
This is to test whether a single model still performs well on multiple scenarios if the total amount of data cannot be increased. For DC the gap increases a little (to -0.70 dB for 2 speakers and -0.80 dB for 3 speakers) and for uPIT it remains similar (+0.06 dB for 2 speakers and -0.18 dB for 3 speakers). This confirms that it is useful to share data between scenarios and only a single model needs to be retained.

\section{conclusions}
\label{sec:conc}
Generally, a model trained on one type of mixture performs suboptimally on another type of mixture. In this paper it was shown that it is useful for a single scenario task to include data from another scenario. Furthermore, we conclude that a single model can be used to cope with different scenarios without significant loss in performance. We hope that other researchers in the MSSS field will consider multi-scenario learning and testing in the feature, instead of being restricted to a single type of mixture.

\section{Acknowledgements}
This work was funded by the SB PhD grant of the Research Foundation Flanders
(FWO) with project number 1S66217N and the KULeuven research grant
GOA/14/005 (CAMETRON).

\vfill\pagebreak

\bibliographystyle{IEEEbib}
\bibliography{refs}

\end{document}